\documentclass[sigconf]{acmart}
\AtBeginDocument{%
  }

\copyrightyear{2025}
\acmYear{2025}
\acmYear{2025}
\setcopyright{cc}
\setcctype{by}
\acmConference[MM '25]{Proceedings of the 33rd ACM International Conference
on Multimedia}{October 27--31, 2025}{Dublin, Ireland}
\acmBooktitle{Proceedings of the 33rd ACM International Conference on
Multimedia (MM '25), October 27--31, 2025, Dublin,
Ireland}\acmDOI{10.1145/3746027.3755728}
\acmISBN{979-8-4007-2035-2/2025/10}

\usepackage{makecell}
\usepackage{multirow}
\usepackage{xcolor}
\usepackage{colortbl}
\newcommand{\ie}{\textit{i.e.}}
\newcommand{\eg}{\textit{e.g.}}

\begin{document}

\title{From Pixels to Tokens: Revisiting Object Hallucinations in Large Vision-Language Models}

\author{Yuying Shang}
\authornote{Both authors contributed equally to this research.}
\email{shangyuying21@mails.ucas.ac.cn}
\orcid{0009-0006-8052-2918}
\affiliation{%
  \institution{University of Chinese Academy of Sciences}
  \city{Beijing}
  \country{China}
}

\author{Xinyi Zeng}
\authornotemark[1]
\affiliation{%
  \institution{Dept. of Comp. Sci. and Tech., Institute for AI, Tsinghua University}
  \city{Beijing}
  \country{China}
}

\author{Yutao Zhu}
\authornotemark[1]
\affiliation{%
  \institution{Gaoling School of Artificial Intelligence, Renmin University of China}
  \city{Beijing}
  \country{China}
}

\author{Xiao Yang}
\affiliation{%
  \institution{Dept. of Comp. Sci. and Tech., Institute for AI, Tsinghua University}
  \city{Beijing}
  \country{China}
}

\author{Zhengwei Fang}
\affiliation{%
  \institution{Dept. of Comp. Sci. and Tech., Institute for AI, Tsinghua University}
  \city{Beijing}
  \country{China}
}

\author{Jingyuan Zhang}
\affiliation{%
  \institution{Kuaishou Technology Inc.}
  \city{Beijing}
  \country{China}
}

\author{Jiawei Chen}
\affiliation{%
  \institution{Shanghai Key Laboratory of Multi. Info. Processing, East China Normal University}
  \city{Shanghai}
  \country{China}
}

\author{Zinan Liu}
\affiliation{%
  \institution{University of Chinese Academy of Sciences}
  \city{Beijing}
  \country{China}
}

\author{Yu Tian}
\authornote{Corresponding author.}
\email{tianyu1810613@gmail.com}
\affiliation{%
  \institution{Dept. of Comp. Sci. and Tech., Institute for AI, Tsinghua University}
  \city{Beijing}
  \country{China}
}

\renewcommand{\shortauthors}{Yuying Shang et al.}
%% No italics, no superscripts
%% Use footnote or author note to identify equal contribution and/or contact author info

\begin{abstract}
Hallucination in large vision-language models (LVLMs) is a significant challenge, \ie, generating objects that are not present in the visual input, which significantly compromises the reliability of models. Recent studies often attribute hallucinations to a lack of visual understanding, yet ignore a more fundamental issue: the model's inability to effectively extract or decouple visual features. In this paper, we revisit the hallucinations in LVLMs from an architectural perspective, investigating whether the primary cause lies in the visual encoder (feature extraction) or the modal alignment module (feature decoupling). Motivated by our preliminary findings, we propose a parameter-efficient fine-tuning strategy, PATCH, to mitigate hallucinations in LVLMs. This plug-and-play method can be integrated into various LVLMs, leveraging adaptive virtual tokens to extract object features from bounding boxes, thereby addressing hallucinations stemming from inadequate feature decoupling. PATCH achieves state-of-the-art performance across multiple multi-modal hallucination datasets and demonstrates significant improvements in general capabilities. We hope this work provides deeper insights into the underlying causes of hallucinations in LVLMs, fostering further advancements and innovation in this field. The code will be available at https://github.com/YuyingShang/PATCH.
\end{abstract}





\maketitle

\section{Introduction}
Large vision-language models (LVLMs) have achieved remarkable performance across a broad range of tasks, even surpassing human capabilities in certain scenarios~\citep{xu2023lvlm,li2023blip,10445007}. However, their practical applications are often hindered by multi-modal hallucinations, where models generate factually incorrect or entirely fictitious outputs when interpreting visual features. Recently, various methods have been proposed to address hallucinations in LVLMs, focusing on aspects such as data distribution~\citep{Yu_2024_CVPR,jiang2024hallucination}, training scheme~\citep{zhao2023beyond,tong2024eyes}, and decoding strategy~\citep{zhang2024debiasing,yang2024pensieve}. Despite these advancements, a fundamental question remains unexplored: \textbf{What is the primary cause of multi-modal hallucinations?}

To better address the problem, we start by exploring the intrinsic sources of hallucinations in LVLMs. We hypothesize two potential factors: (1) insufficient extraction of visual features and (2) inadequate decoupling of these features during multi-modal integration. To validate this, we revisit hallucinations in LVLMs from an architectural perspective, focusing on two key components: the visual encoder (responsible for feature extraction) and the modal alignment module (responsible for feature decoupling). First, we assess the role of the visual encoder by combining it with a pre-trained object detection head to perform object recognition on a hallucination dataset. The detected results are then compared with the 
direct inference outputs of the LVLM on the same dataset. 
Our analysis reveals that the primary source of object hallucinations lies in insufficient cross-modal alignment at the modal alignment module, rather than limitations in the encoding capacity of the visual encoder. Subsequently, we augment the LVLM's input sequence with object-related information, demonstrating that the model's resistance to hallucinations is effectively improved.

Motivated by the preliminary findings, we propose a novel fine-tuning strategy to mitigate hallucinations in LVLMs named \textbf{PATCH} (\textbf{P}luggable virtu\textbf{A}l \textbf{T}okens for obje\textbf{C}t \textbf{H}allucinations). PATCH is designed to enable LVLMs to better leverage object-related information, aligning visual and textual features in the semantic space to mitigate object hallucinations. Concretely, PATCH introduces trainable and pluggable virtual tokens between image features and enhanced prompt texts, bridging the gap between the encoded image features and input-augmented texts with minimal parameter tuning. During inference, the fine-tuned virtual token embeddings are added to the original vocabulary, making PATCH a plug-and-play method that is adaptable to various application scenarios.

To validate the effectiveness and generalization of our method, we conduct experiments on three publicly available multi-modal hallucination evaluation datasets across three LVLMs. Notably, when enhancing the LLaVA-v1.5~\citep{liu2024improved}, MiniGPT-4~\citep{zhu2023minigpt}, and MiniGPT-v2~\citep{chen2023minigpt} with PATCH, the accuracy scores on the POPE~\citep{li2023evaluating} dataset have surged from 85.17\% to 90.20\% (an absolute improvement of 5.03\%), 57.67\% to 88.13\% (an absolute improvement of 30.46\%), and 83.33\% to 90.03\% (an absolute improvement of 6.70\%).

Our experiments and methodology have provided an insightful exploration of the fundamental causes of multi-modal hallucinations from a new perspective, offering new ideas for solving multi-modal hallucinations in LVLMs. In summary, our contributions are summarized as follows:
(1) We explore the intrinsic sources of hallucinations in LVLMs, revealing that the inadequate decoupling of textual and visual features during multi-modal integration is the primary factor behind hallucinations in LVLMs.
(2) We propose PATCH, a parameter-efficient fine-tuning strategy, which enables LVLMs to more effectively leverage visual detection information to address hallucination issues.
(3) The effectiveness of PATCH has been validated on three multi-modal hallucination evaluation datasets across three LVLMs, and further exploration of various hallucination types has demonstrated its great potential, especially in handling strong misleading difficulty in questions.

\begin{table}[!t]
\small
    \centering
    \caption{Comparison between the number of samples in detection and inference results, where ``Det.'' denotes Detection and ``Inf.'' denotes Inference.}
    \vspace{-0.5em}
    \begin{tabular}{lrr}
        \toprule
        {} & Correct Inf. & Wrong Inf. \\
        \midrule
        Correct Det. & 2,396 & 308 \\
        Wrong Det. & 105 & 191 \\
        \bottomrule
    \end{tabular}
    \vspace{-0.5em}
    \label{tab:detect_results}
\end{table}

\section{Related Work}
\subsection{Large Vision-Language Models}

Large Vision-Language Models (LVLMs) represent a significant advancement in multi-modal artificial intelligence, combining visual perception and natural language understanding to process and generate contextual information. The typical LVLM architecture primarily consists of a large language model~\citep{touvron2023llama,touvron2023llama2} for text processing and a visual encoder~\citep{radford2021learning,fang2023eva,sun2023eva} for feature extraction. By incorporating various cross-modal alignment mechanisms, such as MLP~\citep{tolstikhin2021mlp}, attention mechanism~\citep{alayrac2022flamingo}, and Q-Former~\citep{NEURIPS2023_9a6a435e,li2023blip}, LVLMs achieve seamless integration between visual and textual embeddings. In recent years, LVLMs have demonstrated remarkable performance across multiple tasks, particularly in image captioning~\citep{stefanini2022show,chen2023sharegpt4v}, visual question answering~\citep{xu2024lvlm,Li_2024_CVPR}, and visual reasoning and generation~\citep{chen2024spatialvlm}. Models like InstructBLIP~\citep{NEURIPS2023_9a6a435e}, MiniGPT~\citep{zhu2023minigpt,chen2023minigpt}, and LLaVA~\citep{liu2024visual} leverage pretrained cross-modal feature alignment and the instruction fine-tuning strategy to enhance the vision-language representation synergy, improving the model’s ability in complex instruction comprehension. However, empirical studies reveal that current LVLMs suffer from hallucination issues, frequently generating descriptions containing objects absent from the input visual content~\citep{bai2024hallucination,liu2024survey}. In this paper, we conduct systematic experimental analysis to explore the underlying cause of hallucination in LVLMs, proposing a novel fine-tuning strategy to alleviate the object hallucination issue in LVLMs.

\begin{table}
\small
    \centering
    \caption{Inference results of MiniGPT-v2 using Prompt$_1$ and Prompt$_2$ on the POPE dataset.}
    \begin{tabular}{lccccc}
        \toprule
        {} & Accuracy & F1 \\ % & Yes Ratio \\
        \midrule
        Prompt$_1$ & 0.833 & 0.822 \\%& 0.439 \\
        Prompt$_2$ & \textbf{0.888} & \textbf{0.888} \\ %& \textbf{0.494} \\
        \bottomrule
    \end{tabular}
    \label{tab:inf_results}
\end{table}

\subsection{Object Hallucinations in LVLMs}

Hallucination refers to the generation of irrelevant, factually inconsistent, or semantically meaningless content relative to the given context~\citep{liu2024exploring,chen2025exploring}. For LVLMs, this phenomenon primarily manifests as the misalignment between the generated textual descriptions and the corresponding visual input. Recent research has increasingly focused on addressing this critical challenge to improve LVLM reliability.~\citet{tong2024eyes} explored the gap between the visual embedding space of CLIP and vision-only self-supervised learning, identifying that the insufficient visual comprehension is the primary factor for multi-modal hallucination. \citet{jiang2024hallucination} further analyzed the representation distribution in LVLMs for both text and visual tokens, revealing the significant misalignment in cross-modal representations. \citet{zhou2023analyzing} examined the hallucinated textual outputs of LVLMs, finding that the object hallucination issue is closely correlated with the inherent uncertainties during the beam search processes, suggesting that both data distribution and decoding methods contribute to the hallucination issue. 

Building upon these findings, current hallucination mitigation approaches primarily focus on optimizing the data distribution~\citep{Yu_2024_CVPR,jiang2024hallucination}, improving the training paradigm~\citep{zhao2023beyond,huang2024opera,chen2024autobreach}, employing human feedback~\citep{sun2023aligning,jing2024fgaif}, and designing novel decoding strategies~\citep{zhang2024debiasing,yang2024pensieve}. However, these methods frequently encounter limitations, including excessive computational resources and data requirements, and the reliance on manually engineered fine-tuning prompts. In contrast to existing approaches, we propose PATCH, a novel parameter-efficient tuning strategy, which revisits LVLM hallucinations through architectural perspective, effectively mitigating multi-modal hallucination across diverse application scenarios.

\section{Analysis of Object Hallucinations in LVLMs}
We first hypothesize two potential sources of multi-modal object hallucinations in LVLMs. Then, we design and conduct a series of experiments to quantify the impact of these sources. Finally, we thoroughly discuss and analyze the experimental results, offering a solution to mitigate the hallucination issue.

\subsection{Potential Sources of Hallucination}
The architecture of LVLMs typically consists of three components: a visual encoder, a visual projection layer, and a large language model (LLM). The visual encoder is responsible for extracting image features, while the visual projection layer decouples and aligns these features with the semantic space of the LLM. The LLM is responsible for interpreting both the text and image features to generate appropriate responses. Since both the visual encoder and the visual projection layer process image information, they are inherently prone to introducing object hallucinations. Therefore, we hypothesize that multi-modal object hallucinations in LVLMs stem from two main sources: (1) \textbf{Insufficient extraction of visual features.} The visual encoder may fail to capture critical details or misinterpret objects within the image, leading to inaccurate or incomplete visual representations. (2) \textbf{Inadequate decoupling of visual features.} Even when the visual encoder extracts accurate features, the visual projection layer may struggle to correctly align them with the corresponding textual embeddings in the joint semantic space. This misalignment hinders the LLM's ability to integrate and interpret visual information, resulting in hallucinatory outputs. To investigate the root source of the hallucinations, we design and conduct the following experiments.

\begin{figure*}[!t]
    \centering
    \includegraphics[width=.75\textwidth]{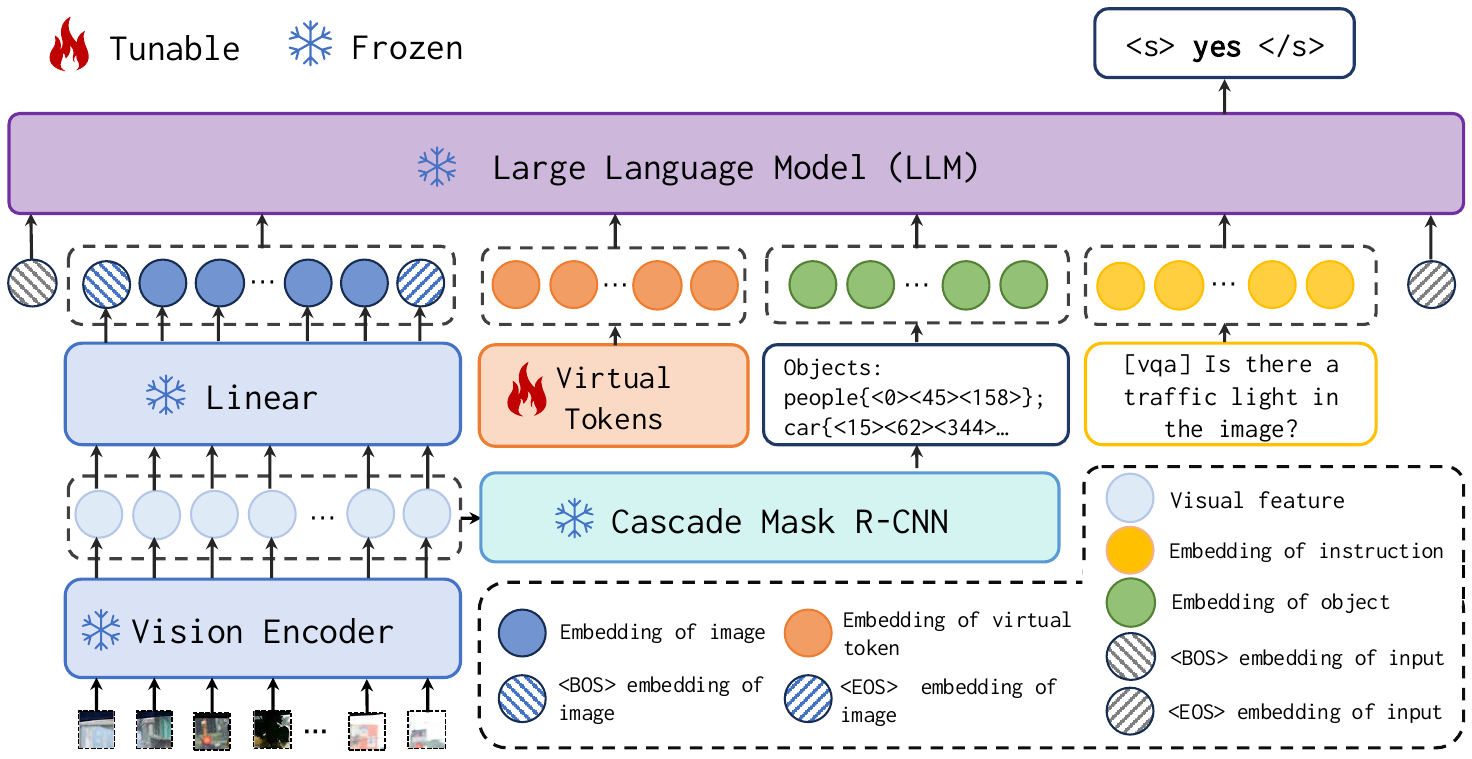}
    \caption{The architecture of LVLMs with PATCH (taking MiniGPT-v2 as an example) where the visual encoder, linear projection layer, and the LLM remain frozen during the fine-tuning phase. The only updated component is the parameters of the virtual tokens. A frozen pre-trained Cascade Mask R-CNN head is adopted to obtain the object-related information in the test images.}
    \label{fig:model}
\end{figure*}

\subsection{Empirical Analysis}\label{sec:pre}

\textbf{Object Hallucination Dataset.}\quad Our experiments are conducted on the POPE~\citep{li2023evaluating} dataset, a dedicated dataset for assessing object hallucination in vision-language models. The dataset consists of 3,000 samples that are highly related to the presence condition of objects within an image. The evaluation is formulated as a binary classification task, where the model is prompted to respond with ``yes'' or ``no'' to questions such as, ``\textit{ Is there a bicycle in the image}? '' Given the binary labels, we can directly assess whether the model is hallucinating or not without applying complex parsing rules. A model is considered to be hallucinating if it answers ``yes'' for an object not present in the image, or ``no'' for an object that is present. 

% Examples of POPE are illustrated in Appendix \textcolor{red}{A}.

\textbf{Experimental Setup.}\quad We conduct a preliminary experiment by MiniGPT-v2~\citep{chen2023minigpt}, which employs a Vision Transformer (ViT)~\citep{dosovitskiy2020image} as its visual encoder.
% , initialized by pre-trained parameters from EVA~\citep{fang2023eva}. 
To validate the image encoding capability of MiniGPT-v2's visual module, we combine it with a pre-trained Cascade Mask R-CNN~\citep{8917599} head, which serves as the object recognition model. We design a prompt, denoted as Prompt$_1$, to evaluate the zero-shot inference performance of MiniGPT-v2, formulated as: 
{\texttt{<Img>[Image]</Img>[vqa][Question]}}, where \texttt{[Image]} and  \texttt{[Question]} are placeholders for the input image and question, while \texttt{[vqa]} serves as MiniGPT-v2's task identifier.
To quantify the alignment between detection and inference, we count the number of correctly and incorrectly identified samples inferred by the model, as shown in Table~\ref{tab:detect_results}. Ideally, the correct object detections should correspond to accurate model outcomes, while incorrect detections should result in faulty inferences. Any deviation or inconsistency between the detected objects and the model’s final inference results is regarded as an instance of object hallucination.

\textbf{Analysis.}\quad As shown in Table~\ref{tab:detect_results}, 413 samples (308 correct detections with incorrect inferences and 105 incorrect detections) exhibit object hallucinations in MiniGPT-v2, accounting for 13.77\% of the total dataset. Notably, 74.58\% of these cases arise when object detection is accurate while the model’s inference is incorrect, indicating this as the predominant failure mode. 

This finding suggests that while MiniGPT-v2’s visual encoder is able to extract accurate image features, its visual projection module struggles to seamlessly bridge the encoded features with the LLM's semantic space. The fundamental misalignment between visual feature encoding and the LLM's semantic interpretation constitutes a primary factor contributing to multi-modal object hallucinations.
% in LVLMS.

\textbf{Potential Solutions.}\quad In many computer vision tasks~\citep{hafiz2020survey,Deng_2021_ICCV}, objects are typically characterized by their categories and bounding boxes. We hypothesize that incorporating object-related information directly into the prompt texts can mitigate the misalignment between visual features and the LLM's semantic space, improving the model's understanding of objects in images. To test this hypothesis, we conduct an additional experiment using a modified prompt, denoted as Prompt$_2$: {\texttt{<Img>[Image]</Img>Objects: \texttt{[Object][vqa][Question]}}. The object-related information is inserted at the \texttt{[Object]} position, formatted as \texttt{category\{<x1><x2>\\}\texttt{<x3><x4>\}}, where each detected object category is concatenated with its corresponding bounding boxes. This inclusion of object recognition details is intended to enhance the LLM's ability to comprehend and process the object information within images. Table~\ref{tab:inf_results} shows the results of MiniGPT-v2 based on Prompt$_1$ and Prompt$_2$. It is evident that incorporating accurate detection information significantly improves the model's performance on questions related to object existence, thus enhancing its image interpretation capability and effectively reducing object hallucinations.

\textbf{Motivation.}\quad While our preliminary experiments identify the potential causes of object hallucinations and suggest a possible solution, we observe that directly incorporating detection information into the input may introduce unnecessary details, particularly when detected objects are weakly correlated with the given question. This redundancy may interfere with the model's reasoning process, hindering its ability to effectively utilize relevant information. To address this issue, we propose the PATCH tuning strategy, which employs trainable and pluggable virtual tokens to help the LLM selectively optimizes the use of detection information. These virtual tokens serve as a bridge, allowing LVLMs to focus on task-relevant image features, improving alignment between visual and textual representations within the semantic space.

\section{Methodology}
In this section, we first introduce the task formulation and then provide a comprehensive description of our method, including its theoretical foundations and practical implementation.

\subsection{Problem Formulation}
The LVLMs aim to generate proper text responses to multi-modal inputs by integrating visual and textual information. The standard approach involves extracting visual features through a visual encoder, mapping these features into the linguistic semantic space via a projection layer for cross-modal fusion and alignment. The fused representations are then decoded by an autoregressive language model to generate the final response. Formally, given an input image $I$ and a corresponding question or text prompt $Q$, the generated answer sequence $Y$ is calculated as:
\begin{align}
p(Y) = \prod\limits_{t=1}^{T} p_{\theta}(y_t \mid I, Q, y_{<t}),
\end{align}
where $y_{<t}$ denotes the sequence of tokens prior to the current token $y_t$ at step $t$, and $\theta$ represents the parameters of the LVLM. Based on our preliminary analysis, we observe that incorporating object categories $C$ and their corresponding bounding boxes $B$ can enhance the generation process. Therefore, the generation process can be reformulated as:
\begin{align}
p(Y) = \prod\limits_{t=1}^{T} p_{\theta}(y_t \mid I, D, Q, y_{<t}),
\end{align}
where $D=[C(I);B(I)]$. In the object hallucination recognition task, the input question $Q$ is designed to assess the presence of hallucinated objects. The task aims to generate the correct answer $Y$ by evaluating whether a given object appears in the image $I$.

\subsection{PATCH}
The proposed fine-tuning strategy aims to mitigate object hallucinations in LVLMs by introducing trainable virtual tokens that leverage additional object-related information. The architecture of our method (taking MiniGPT-v2 as an example) is illustrated in Figure~\ref{fig:model}. Inspired by~\citep{zhu2024one}, we insert a set of $n$ virtual tokens $T = [t_1, t_2,\ldots, t_n]$ between the image features and the detection information $D$. These token embeddings are optimized during training, with parameters $\delta \in \mathcal{R}^{n\times d}$, where $d$ is the embedding dimension of the LVLM. The generation process of LVLMs, augmented by these virtual tokens, is formulated as:
\begin{align}
p(Y) = \prod\limits_{t=1}^{T} p_{\delta,\theta}(y_t \mid I, [t_1, t_2, \ldots, t_n], D, Q, y_{<t}).
\end{align}

To reduce the computing resources, all LVLM parameters $\theta$ are frozen during fine-tuning, except for the newly introduced parameters $\delta$ of virtual tokens. For instance, with 20 additional virtual tokens, only $20 \times 4,096 = 0.08$M parameters are trainable, accounting for merely 0.0012\% of the total model parameters. This approach achieves substantial computational efficiency while preserving its optimization capability for object hallucination mitigation. Comprehensive experimental analysis is demonstrated in Section~\ref{main results}.

During the inference phase, we extend the model's vocabulary by incorporating several special tokens (\eg, \texttt{[ref1]}, \texttt{[ref2]}, $\ldots$, \texttt{[refn]}) whose embeddings are initialized by the fine-tuned virtual token embeddings. This makes PATCH a plug-and-play method that can be dynamically adjusted based on application requirements. Specifically, when object-related information is included in the users' input, virtual tokens can be added before the detection results, helping to mitigate object hallucinations in LVLMs. Conversely, when no additional detection information is provided, the LVLM can revert to processing the input using its standard capabilities without PATCH involvement. PATCH strengthens image comprehension by optimizing the alignment between visual features and textual semantics while preserving the model’s inherent capabilities. This adaptability is particularly valuable in practical applications, as LVLMs are commonly deployed across diverse downstream tasks. By narrowing the representational gap between modalities, PATCH improves cross-modal feature fusion, particularly for tasks that benefit from enriched detection prompts, ultimately reducing object hallucinations and enhancing the overall performance of LVLMs.

\section{Experiments}
\subsection{Datasets}

To verify the effectiveness of our method, we conduct experiments on two publicly available multi-modal hallucination evaluation datasets. (1) \textbf{POPE} dataset~\citep{li2023evaluating} is specifically designed for evaluating object hallucinations in LVLMs, which comprises 3,000 adversarially constructed samples from A-OKVQA and MSCOCO, respectively. The A-OKVQA portion is treated as the training set, while the MSCOCO portion is for testing. (2) \textbf{PhD} dataset~\citep{liu2024phd} is a newly introduced benchmark for multi-modal hallucination evaluation. We conduct experiments on its $\text{v}_1$ version across five task types that are highly related to objects, including Object Recognition, Attribute Recognition, Counting, Positional Reasoning, and Sentiment Analysis. 80\% of the data are randomly selected for training, while the rest is for testing. 

% Further details can be found in Appendix \textcolor{red}{A}.

\subsection{Implementation Details}\label{Implementation Details}

The LLaVA-v1.5, MiniGPT-4, and MiniGPT-v2 are adopted as the backbone LVLMs, initializing them with their default parameter configurations. Given the advanced capabilities of MiniGPT-v2, most of our experiments are conducted by this model. Accuracy, Precision, Recall, and F1 score are employed as evaluation metrics. The cosine scheduler~\citep{loshchilov2016sgdr} is utilized to adjust the learning rate and the LLaMA-2-7B-chat~\citep{touvron2023llama} is employed as the language model. 

All backbone models are fine-tuned over 20$ \times d$ trainable parameters by default, where $d$ represents the dimensionality of the hidden layers. As demonstrated in our preliminary experiments (Section~\ref{sec:pre}), incorporating object-related information directly into the input prompt (Prompt$_2$) can also improve performance. We refer to this simple approach as the ``\textbf{Hard Prompt}'' method. All training is performed on a single NVIDIA A100 GPU. 

% Further implementation details can be found in Appendix \textcolor{red}{B}.

\subsection{Baselines}\label{baselines}
We consider nine mainstream LVLMs: (1) \textbf{mPLUG-Owl}~\citep{ye2023mplug} is a two-stage training framework that aligns visual and textual modalities by training a visual knowledge module followed by an abstraction module. (2) \textbf{Multi-modal-GPT}~\citep{gong2023multimodal} utilizes a specially designed input sequence and calculates the loss based on the response and the end token to enhance visual-text alignment. (3) \textbf{InstructBLIP}~\citep{NEURIPS2023_9a6a435e} employs a Q-Former~\citep{li2023blip} mechanism to compress visual features into a fixed number of tokens, which are then concatenated with text tokens. This approach facilitates the comprehension of visual and linguistic information through instruction tuning. (4) \textbf{LLaVA}~\citep{liu2024visual} and (5) \textbf{MiniGPT-4}~\citep{zhu2023minigpt} use a linear projection layer for cross-modal embedding alignment. LLaVA emphasizes fine-tuning for visual-language alignment by specific instructions, while MiniGPT-4 is optimized for generating detailed descriptions from images. (6) \textbf{LLaVA-v1.5}~\citep{liu2024improved} and (7) \textbf{MiniGPT-v2}~\citep{chen2023minigpt} are the improved version of the original model, optimized for better cross-modal understanding. (8) \textbf{Osprey}~\citep{YuanLLTLQZZ24} introduces a mask-based visual extractor to extract features from high-resolution images, enabling fine-grained and open-world visual understanding. (9) \textbf{GLAMM}~\citep{rasheed2024glamm} is the first model capable of generating natural language responses intertwined with corresponding object segmentation masks, enabling in-depth region understanding.

In addition to these LVLMs, we compare our method with three recently released hallucination-solving methods based on LLaVA-v1.5 and MiniGPT-4:
(1) \textbf{HA-DPO}~\citep{zhao2023beyond} reformulates the hallucination problem as a preference optimization task, training the model to consistently favor the accurate responses over hallucinatory ones with the given image-question pair. 
(2) \textbf{Woodpecker}~\citep{yin2023woodpecker} proposes a five-stage framework to locate the hallucinations and verify factuality, introducing a training-free method to correct hallucinations from the generated texts. 
(3) \textbf{HACL}~\citep{jiang2024hallucination} incorporates contrastive learning into the training process by treating hallucinatory text as hard negative examples, encouraging the model to align non-hallucinatory texts more closely with their corresponding images. 

\begin{table}[!t]
\footnotesize
  \begin{center}
  \caption{Performance of LVLMs and hallucination mitigation methods on the POPE dataset. The best results are highlighted in \textbf{bold}. The subscript values indicate the improvement over the backbone models.}
  \resizebox{\columnwidth}{!}{
  \begin{tabular}{lcccc}
    \toprule
    \textbf{Model} & \textbf{Accuracy} & \textbf{Precision} & \textbf{Recall} & \textbf{F1} \\
    \midrule
    mPLUG-Owl & 50.67 & 50.34 & 99.33 & 66.82 \\
    Multi-modal-GPT & 50.00 & 50.00 & \textbf{100.00} & 66.67 \\
    InstructBLIP & 74.37 & 67.67 & 93.33 & 78.45 \\
    LLaVA & 50.77 & 50.39 & 99.87 & 66.98 \\
    % Qwen2-VL-7B & 86.83 & 88.02 & 85.27 & 86.62 \\
    Osprey & 85.33 & 85.43 & 85.20 & 85.31 \\
    GLaMM & 87.30 & 85.03 & 90.53 & 87.70 \\
    \midrule
    LLaVA-v1.5 & 85.17 & 89.93 & 79.20 & 84.23 \\
    \quad + HA-DPO & 81.46$_{\text{[-3.71]}}$ & 77.99$_{\text{[-11.94]}}$ & 87.66$_{\text{[+8.46]}}$ & 82.54$_{\text{[-1.69]}}$ \\
    \quad + HACL & 86.54$_{\text{[+1.37]}}$ & \textbf{93.01}$_{\text{[+3.08]}}$ & 79.52$_{\text{[+0.32]}}$ & 85.73$_{\text{[+1.50]}}$ \\
    \quad + Hard Prompt & 89.93$_{\text{[+4.76]}}$ & 91.14$_{\text{[+1.21]}}$ & 88.47$_{\text{[+9.27]}}$ & 89.78$_{\text{[+5.55]}}$ \\
    \cellcolor[RGB]{235,245,250}{\quad + PATCH (ours)} & \cellcolor[RGB]{235,245,250}{\textbf{90.20}$_{\text{[+5.03]}}$} & \cellcolor[RGB]{235,245,250}{91.13$_{\text{[+1.20]}}$} & \cellcolor[RGB]{235,245,250}{89.07$_{\text{[+9.87]}}$} & \cellcolor[RGB]{235,245,250}{\textbf{90.09}$_{\text{[+5.86]}}$} \\
    \midrule
    MiniGPT-4 & 57.67 & 54.29 & 96.93 & 69.60 \\
    \quad + HA-DPO & 75.66$_{\text{[+17.99]}}$ & 74.36$_{\text{[+20.07]}}$ & 78.33$_{\text{[-18.60]}}$ & 76.29$_{\text{[+6.69]}}$ \\
    \quad + Woodpecker & 82.33$_{\text{[+24.66]}}$ & 83.92$_{\text{[+29.63]}}$ & 80.00$_{\text{[-16.93]}}$ & 81.91$_{\text{[+12.31]}}$ \\
    \quad + HACL & 71.32$_{\text{[+13.65]}}$ & 70.53$_{\text{[+16.24]}}$ & 73.45$_{\text{[-23.48]}}$ & 71.96$_{\text{[+2.36]}}$ \\
    \quad + Hard Prompt & 70.73$_{\text{[+13.06]}}$ & 63.60$_{\text{[+9.31]}}$ & 96.93$_{\text{[+0.00]}}$ & 76.81$_{\text{[+7.21]}}$ \\
    \cellcolor[RGB]{235,245,250}{\quad + PATCH (ours)} & \cellcolor[RGB]{235,245,250}{{88.13$_{\text{[+30.46]}}$}} & \cellcolor[RGB]{235,245,250}{86.99$_{\text{[+32.70]}}$} & \cellcolor[RGB]{235,245,250}{89.67$_{\text{[-7.26]}}$} & \cellcolor[RGB]{235,245,250}{88.31$_{\text{[+18.71]}}$} \\
    \midrule
    MiniGPT-v2 & 83.33 & 88.28 & 76.87 & 82.18 \\
    \quad + Hard Prompt & 88.77$_{\text{[+5.44]}}$ & 88.23$_{\text{[-0.05]}}$ & 89.47$_{\text{[+12.60]}}$ & 88.84$_{\text{[+6.66]}}$ \\
    \cellcolor[RGB]{235,245,250}{\quad + PATCH (ours)} & \cellcolor[RGB]{235,245,250}{{90.03$_{\text{[+6.70]}}$}} & \cellcolor[RGB]{235,245,250}{91.39$_{\text{[+3.11]}}$} & \cellcolor[RGB]{235,245,250}{88.40$_{\text{[+11.53]}}$} & \cellcolor[RGB]{235,245,250}{89.87$_{\text{[+7.69]}}$} \\
  \bottomrule
\end{tabular}}
\label{main result}
\end{center}
\end{table}

\subsection{Experimental Results}\label{main results}

\begin{figure*}[t]
    \centering
    \begin{minipage}[b]{.43\linewidth}
        \includegraphics[width=\linewidth]{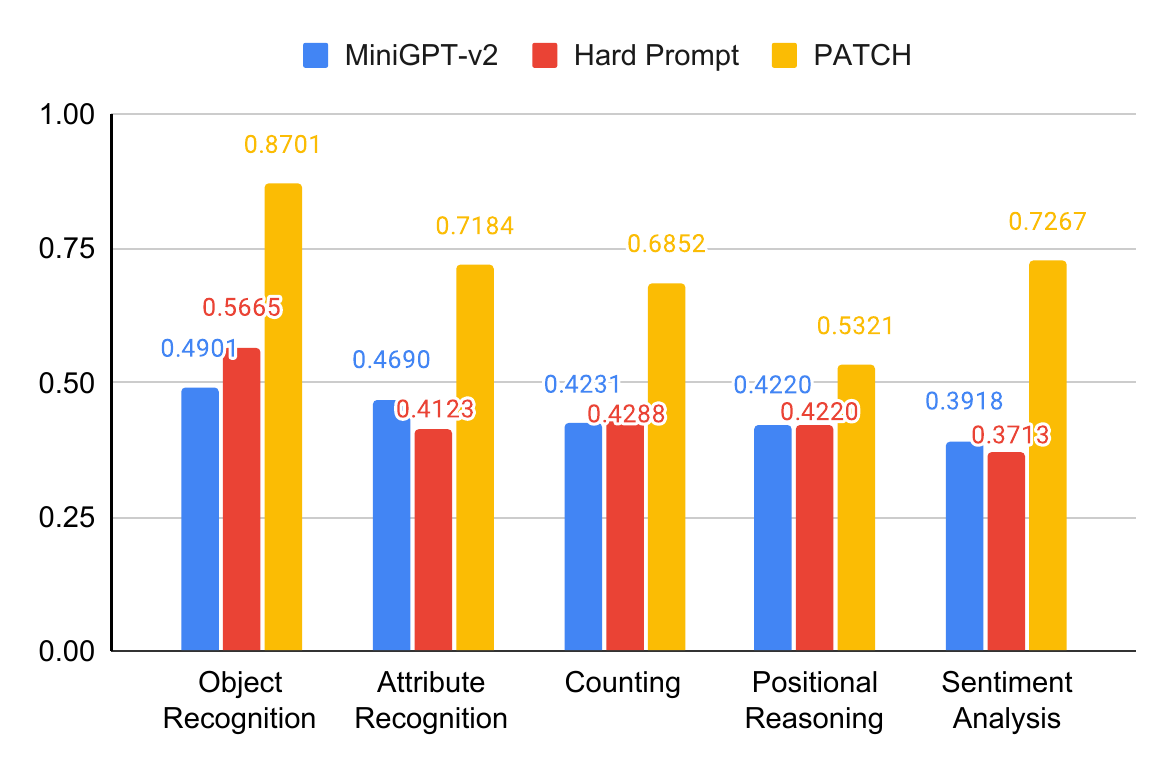}
    \end{minipage}
    \begin{minipage}[b]{.33\linewidth}
        \includegraphics[width=\linewidth]{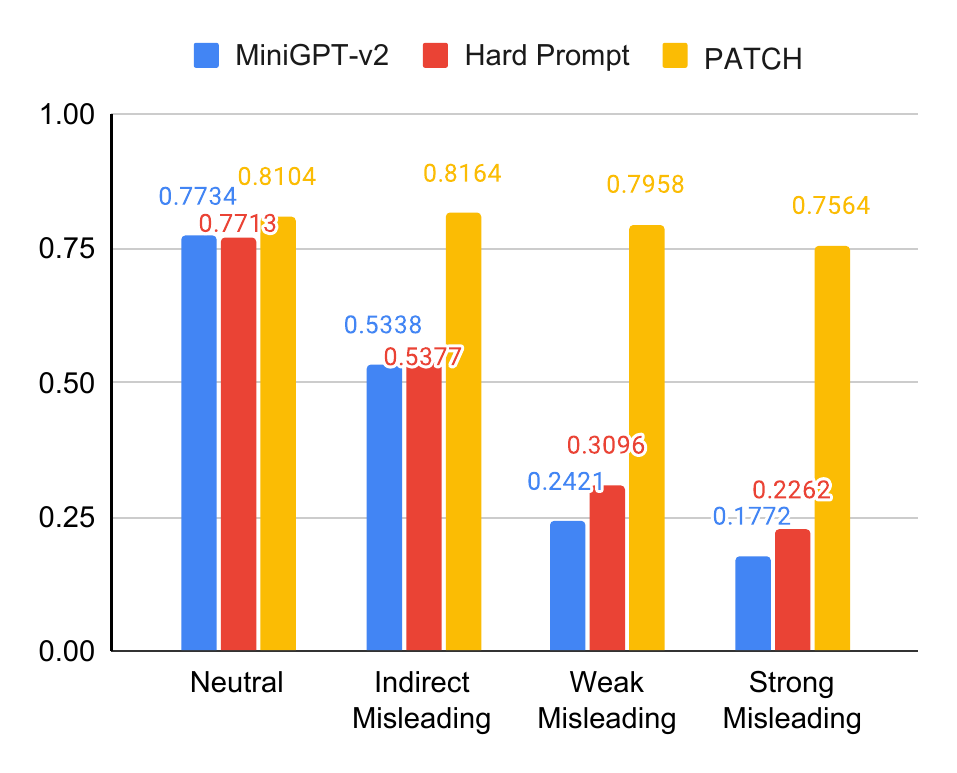}
    \end{minipage}
    \caption{Performance on the PhD dataset across different task types (left) and conflict levels (right).}
    \label{fig:phd}
\end{figure*}

\textbf{Analysis on POPE.}\quad
The experimental results are shown in Table~\ref{main result}. It is clear to see that our method significantly improves the performance of three backbone LVLMs in the object hallucination recognition task. Several observations emerge from the results: (1) Pre-trained LVLMs, such as mPLUG-Owl, Multi-modal-GPT, InstructBLIP, MiniGPT-4, and LLaVA are prone to generate hallucinated content, while their advanced versions (LLaVA-v1.5 and MiniGPT-v2) exhibit better performance. We attribute this to their superior training strategies (\ie, instruction tuning). Models with pixel-level understanding, such as Osprey and GLAMM, perform relatively well on POPE, likely benefiting from the incorporation of extensive object-level instruction data during training. (2) Both Hard Prompt and PATCH improve the backbone performance, confirming that incorporating object-related information aids LVLMs in better interpreting visual features. Specifically, PATCH outperforms Hard Prompt as the parameters of virtual tokens are optimized during fine-tuning, allowing precise alignment between image content and corresponding texts, which effectively mitigates object hallucinations in LVLMs. (3) Compared to three previous approaches designed to alleviate object hallucinations, PATCH achieves notable accuracy improvements of 30.46\%, 5.03\%, and 6.70\% on LLaVA-v1.5, MiniGPT-4 and MiniGPT-v2, respectively, achieving the state-of-the-art performance. Unlike HA-DPO and HACL, which rely on complex optimization techniques, PATCH achieves superior results with minimal parameter tuning. While Woodpecker proposes a training-free solution, its reliance on manually crafted prompts limits its flexibility and scalability. In contrast, PATCH introduces a parameter-efficient fine-tuning strategy via pluggable virtual tokens, enabling the model to adaptively extract valuable information from detection information during the fine-tuning process. 

\begin{table}[!t]
\small
\begin{center}
\caption{The ablation results on object-related information and virtual token position on the POPE dataset, where ``categ.'' denotes categories.}
  \begin{tabular}{lcccc}
    \toprule
    \textbf{Method} & \textbf{Accuracy} & \textbf{Precision} &	\textbf{Recall} & \textbf{F1} \\
    \midrule
    PATCH & \textbf{90.03} & \textbf{91.39} & 88.40 & \textbf{89.87} \\
    \quad \textit{w/o} bbox & 88.27 & 87.37 & 89.47 & 88.41 \\
    \quad \textit{w/o} bbox \& categ. & 82.60 & 84.93 & 79.27 & 82.00 \\
    \quad \textit{w/} Late & 87.60 & 85.79 & \textbf{90.13} & 87.91 \\
  \bottomrule
\end{tabular}
\label{table:detection results}
\end{center}
\end{table}

\textbf{Analysis on PhD.}\quad We compare PATCH with the Hard Prompt method based on MiniGPT-v2 across different task types and conflict levels on the PhD dataset: 
(1) \textbf{Task types}. The accuracy scores of the three variants across five task types are presented on the left side of Figure~\ref{fig:phd}. The results show that using the Hard prompt to inject object-related information can improve performance in object recognition, counting, and positional reasoning tasks. This is consistent with our findings in preliminary experiments, where object-related information enhances the LVLM's understanding of geometric information of objects in images. However, for object attribute recognition and sentiment analysis tasks, the hard prompt brings some negative effects. This may be due to the fact that the detection results do not include information directly relevant to the given questions. Consequently, inserting complex object-related information into the input texts may introduce excessive redundant noise, interfering with the ability of LVLMs to effectively leverage their prior knowledge. Fortunately, PATCH improves performance across all tasks, which clearly demonstrates that the fine-tuned virtual tokens can effectively guide the model in extracting valuable information from detection results as well as mining the prior knowledge within the LLM. 
\begin{table}[!t]
\small
    \centering
    \caption{Experimental results by replacing the object categories with the generic placeholder.}
    \begin{tabular}{lrrrrr}
        \toprule
        \textbf{Method} & \textbf{Accuracy} & \textbf{Precision} & \textbf{Recall} & \textbf{F1}\\
        \midrule
        MiniGPT-v2 & 83.37 & 88.00 & 77.27 & 82.29 \\
        \quad \textit{w/} placeholder & 80.70 & 86.40 & 72.87 & 79.06 \\
        PATCH & \textbf{90.03 }& \textbf{91.39} & \textbf{88.40} & \textbf{89.87 }\\
        \quad \textit{w/} placeholder & 88.73 & 88.89 & 88.53 & 88.71 \\
        \bottomrule
    \end{tabular}
    \label{tab:placeholder}
\end{table}
(2) \textbf{Conflict levels}. On the right side of Figure~\ref{fig:phd}, we present the accuracy scores of the three variants across four conflict levels. The PhD dataset provides three statements per question that are entirely inconsistent with the image content, serving as misleading context to challenge the model during answer generation. The misleading statements are categorized into three types based on their intensity: strong misleading, weak misleading, and indirect misleading, while neutral represents the original question text without added misleading statement. From the results, it is obvious that the three variants perform similarly on neutral questions, indicating that recent pre-trained LVLMs are able to answer questions accurately in the absence of misleading information. However, as misleading statements are introduced into the context, the performance of MiniGPT-v2 and Hard Prompt declines significantly. This suggests that LVLMs struggle to handle questions with misleading texts, and simply adding object-related information as prompts is insufficient to mitigate this issue. In contrast, the proposed PATCH performs remarkably well even on strongly misleading questions, showing that fine-tuning LVLMs with trainable virtual tokens can effectively improve the model's ability to discern the true relationships between textual and visual content. This demonstrates the robust cross-modal information understanding capability of PATCH, showcasing its effectiveness in addressing the object hallucination issue in LVLMs.

\begin{figure}[!t]
    \centering
    \includegraphics[width=.78\columnwidth]{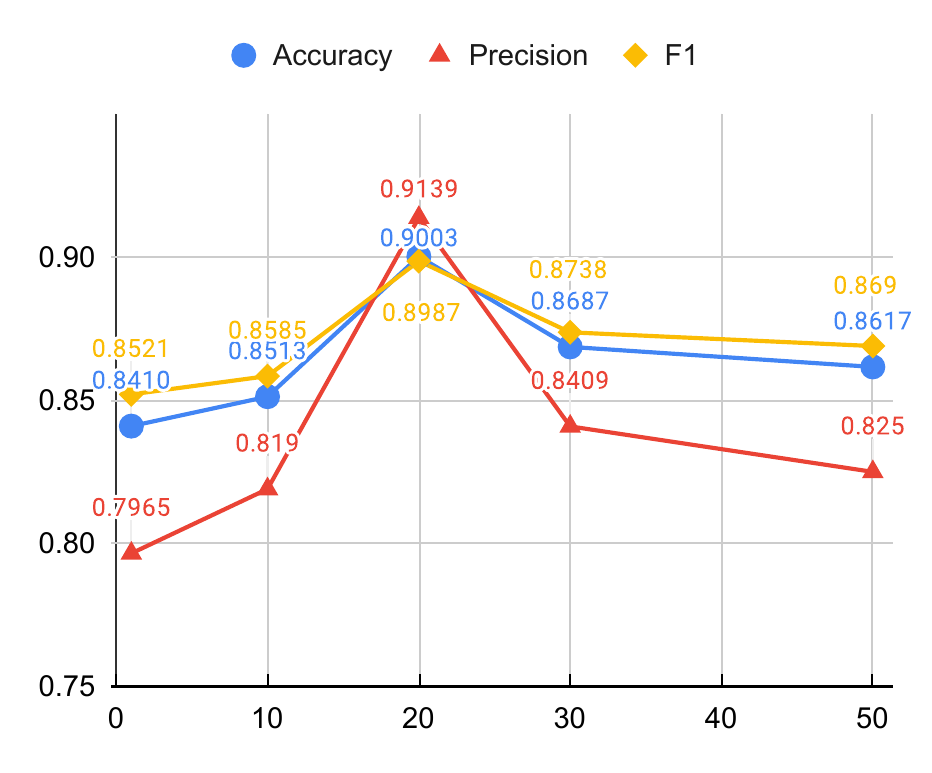}
    \caption{Accuracy, Precision, and F1 score over different token quantities on POPE dataset.}
    \label{fig:token_quantity}
\end{figure}

\subsection{Ablation Study}

\subsubsection{Impact on Object-related Information.} (1) \textbf{Recognition Information.} We explore the impact of various object detection results on PATCH, and the experimental results are shown in Table~\ref{table:detection results}. The ``bboxes'' is denoted as the bounding box information and ``categ.'' represents the object category information. The experimental results show that when complete detection information is provided, PATCH achieves the highest accuracy of 90.03\% on the POPE dataset, demonstrating its effectiveness in enhancing LVLMs' understanding of scenes and objects. When the bounding box information is removed, the accuracy drops by 1.76\%, suggesting that bounding boxes play an important role in object localization and scene comprehension. However, a slight increase in the recall score is observed at the same time, suggesting that without bounding boxes, the model becomes more effective at detecting positive samples. This may be because bounding boxes sometimes cause the model to focus narrowly on specific regions, thereby limiting its understanding of the broader image context. When all object-related information is omitted, leaving only 20 trainable virtual tokens for fine-tuning, the model's performance drops significantly. This indicates that without the additional object-related information, solely fine-tuning virtual tokens is insufficient to achieve consistent semantic alignment between visual and textual features. Consequently, the model exhibits increased uncertainty in object recognition, underscoring the importance of incorporating rich object-related information to mitigate hallucination in LVLMs. (2) \textbf{Impact of Token Position.}\label{Token Position} In our method, virtual tokens are added between the detection results and the question, following the format: ``\texttt{[Image][VirtualTokens][Object][Question]}''. To investigate the effect of virtual token positioning, we consider an alternative format: ``\texttt{[Image][Object][VirtualTokens][Question]}'', referred to as ``PATCH \textit{w/} Late''. The comparison result is shown in Table~\ref{table:detection results}, which indicates that the placement of the virtual tokens significantly impacts model performance. Positioning the virtual tokens between the object-related information and the image features enables the LVLM to better interpret the detection information. In contrast, placing the virtual tokens after the object-related information causes the LVLM to focus excessively on the detection text, while neglecting critical contextual cues from the overall image features, resulting in a performance decline. (3) \textbf{Impact of Localization Information.} Table~\ref{tab:placeholder} exhibits the ablation experimental results by replacing the object categories with a generic placeholder ``\texttt{object\{idx\}}'' to validate the effect of the localization information, where the object-related information is reformatted as \texttt{object\{idx\}\{<x1><x2><x3><x4>\}}. The experimental results demonstrate that merely incorporating object location information into LVLMs for reasoning fails to effectively mitigate hallucinations, likely due to the ambiguous referential details in the additional textual inputs. In contrast, when only location information is utilized for PATCH fine-tuning, hallucinations are partially mitigated. This may be attributed to the fact that the LVLM learns to leverage virtual tokens to comprehend precise object localization information during the fine-tuning process, thereby enhancing image interpretation and ensuring faithful responses to the input image.

\begin{table}[!t]
\small
\begin{center}
\caption{Ablation experimental results over different
token quantities on POPE dataset based on MiniGPT-v2.}
  \begin{tabular}{lcccc}
    \toprule
      {} & \textbf{Accuracy} & \textbf{Precision} &	\textbf{Recall} & \textbf{F1} \\
    \midrule
    MiniGPT-v2 & \textbf{83.37} & \textbf{88.00} & 77.27 & \textbf{82.29} \\
    \quad + 20 tokens & 82.60 & 84.93 & 79.27 & 82.00 \\
    \quad + 30 tokens & 82.27 & 82.97 & 81.20 & 82.08 \\
    \quad + 50 tokens & 81.40 & 78.72 & 86.07 & 82.23 \\
    \quad + 100 tokens & 77.67 & 73.29 & \textbf{87.07} & 79.59 \\
  \bottomrule
\end{tabular}
\label{table:token number}
\end{center}
\end{table}

\begin{table}[!t]
\small
\begin{center}
\caption{Ablation experimental results on different virtual token initializations on POPE dataset.}
\begin{tabular}{lcccc}
\toprule
\textbf{Initialization} &  \textbf{Accuracy} & \textbf{Precision} & \textbf{Recall} & \textbf{F1} \\
\midrule
Random & 86.77 & 84.28 & 90.40 & 87.23 \\
$T_1$ & 88.83 & \textbf{93.76} & 83.20 & 88.17 \\
$T_2$ & \textbf{90.03} & 91.39 & \textbf{88.40} & \textbf{89.87} \\
\bottomrule
\end{tabular}
\label{tab:token initial}
\end{center}
\end{table}

\begin{figure}[!t]
    \centering
    \includegraphics[width=.84\columnwidth]{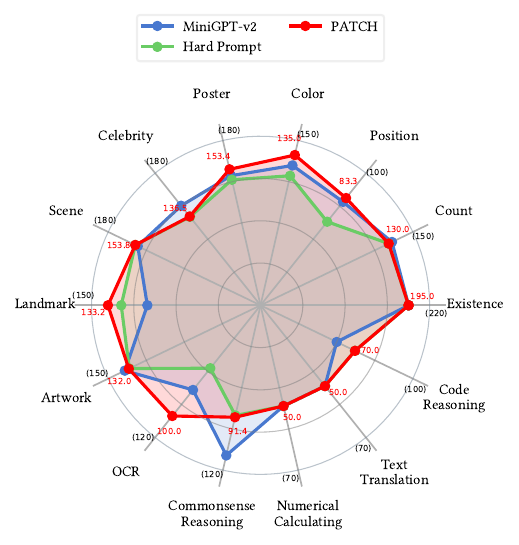}
    \caption{Performance of MiniGPT-v2, Hard Prompt and PATCH on MME benchmark across 14 subtasks.}
    \label{fig:MME}
\end{figure}

\subsubsection{Impact on Token Quantities.} We first evaluate the performance of PATCH across different quantities of virtual tokens. As illustrated in Figure~\ref{fig:token_quantity}, the metric scores initially improve as the number of virtual tokens increases, reaching optimal accuracy with 20 tokens, which yields a 6.70\% improvement over the baseline configuration. However, further increasing the number of tokens results in a noticeable performance decline. This suggests that an excessive number of tokens may introduce redundant information, diminishing the model’s ability to accurately focus on essential object-related information, and ultimately leading to a significant degradation in overall performance. Then, we further explore the impact of virtual token quantities when object-related information is explicitly excluded based on MiniGPT-v2. The experimental results are shown in Table~\ref{table:token number}, which clearly demonstrate the critical role of the incorporation of the detection results. These findings suggest that simply increasing the number of trainable prompting tokens is insufficient to mitigate hallucinations in LVLMs. Moreover, an increased number of virtual tokens raises the fine-tuning costs, leading to a gradual decline in performance when the learning rate and number of training epochs remain fixed. 

\subsubsection{Impact on the Initialization Strategy of Virtual Tokens.} To investigate the effect on the initialization format of virtual tokens, we introduced two distinct strategies: (1) \textbf{Random Initialization} and (2) \textbf{Text-based Initialization} which consists of two instruction texts. The template of $T_1$ refers to ``According to the previous object detection results, please answer the following question'', while $T_2$ refers to ``According to the previous object detection results, please answer the following question with `yes' or `no' ''. The text encoder of the LVLM is leveraged to encode the input prompt, which is configured with a maximum sequence length of $n$. Sequences exceeding this limit are truncated, while shorter sequences are zero-padded to ensure dimensional consistency across all inputs. The experimental results of PATCH are shown in Table~\ref{tab:token initial}, which indicates that random initialization leads to a noticeable decrease in overall model performance. In contrast, initializing tokens with explicit text-based prompts consistently improves all evaluation metrics. Given that the responses in the POPE dataset are constrained to ``yes'' or ``no'', aligning virtual token initialization with the answer distribution enhances the LVLM's ability to effectively utilize virtual tokens for extracting relevant object-related information, resulting in more reliable and standardized response generation.

\subsection{Further Analysis}
% In this section, we conduct a more in-depth analysis of the generalization capability of PATCH.

\textbf{Performance on MME Evaluation Benchmark.}\quad We conduct an experiment on the multi-modal evaluation benchmark, MME~\citep{abs-2306-13394}, to assess PATCH's effectiveness in general-domain tasks. Specifically, PATCH is initialized with parameters pre-trained on the POPE (AOKVQA-version) dataset. Table~\ref{table:MME overall} and Figure~\ref{fig:MME} summarize the performance across the Perception and Cognition tasks in MME. Experimental results demonstrate that PATCH outperforms MiniGPT-v2 by 34.55 points in the Perception task, with particularly strong performance in Position, Color, Poster, Scene, Landmark, and OCR tasks. This improvement is primarily attributed to PATCH’s enhanced alignment with image contents. However, performance declines in Count, Celebrity, and Artwork tasks, which is likely due to the detector errors and the model’s limited understanding of celebrity-related information. These limitations could potentially be mitigated through improved object detectors and retrieval-augmented generation techniques. Despite these improvements, PATCH exhibits an 8.21-point performance decrease on the Cognition task. Since our method has only been fine-tuned on the POPE (AOKVQA-version) dataset, we believe that additional fine-tuning on larger-scale datasets could further enhance its performance. 

% (Relevant analysis can be found in Appendix \textcolor{red}{D}).

\begin{table}[!t]
\small
\centering
\caption{Overall performance on MME benchmark in Perception and Cognition tasks.}
\begin{tabular}{lcc}
\toprule
\textbf{Method} & \textbf{Perception} & \textbf{Cognition}\\
\midrule
MiniGPT-v2 & 1317.45 & \textbf{269.64} \\
\quad + Hard Prompt & 1290.96 & 260.71 \\
\quad + PATCH & \textbf{1352.2} & 261.43\\
\bottomrule
\end{tabular}
\label{table:MME overall}
\end{table}

\begin{table}[!t]
\begin{center}
\small
\caption{Zero-shot experimental results of PATCH on the full PhD dataset using parameters pre-trained on POPE dataset.}
  \begin{tabular}{lcccc}
    \toprule
      \textbf{Method} & \textbf{Accuracy} & \textbf{Precision} &	\textbf{Recall} & \textbf{F1} \\
    \midrule
    MiniGPT-v2 & 47.92 & 52.46 & 55.48 & 53.93 \\
    \quad + Hard Prompt & 49.48 & 53.20 & 66.86 & 59.25 \\
    \quad + PATCH & \textbf{52.84} & \textbf{55.91} & \textbf{66.90} & \textbf{60.91} \\
  \bottomrule
\end{tabular}
\label{table:transfer experiment}
\end{center}
\end{table}

% We conduct a zero-shot experiment where

\textbf{Zero-shot Experiment.} \quad PATCH, initialized with parameters fine-tuned on the POPE (AOKVQA-version) dataset, is evaluated on the full PhD dataset. To ensure the validity, we have thoroughly verified that there is no distribution overlap between the two datasets. The experimental results are shown in Table~\ref{table:transfer experiment}, which demonstrate a 4.92\% improvement in accuracy over the baseline model. This observation further validates PATCH's plug-and-play capability, as it can be directly transferred to a new dataset without additional fine-tuning while still delivering significant performance gains.

\section{Conclusion}
In this paper, we revisit hallucinations in LVLMs from an architectural perspective, hypothesizing two potential causes: (1) insufficient extraction of visual features
and (2) inadequate decoupling of visual features. Based on our preliminary experimental results, we identify that the primary cause lies in insufficient cross-modal alignment rather than deficiencies in the visual encoding process. Motivated by this insight, we propose PATCH, a novel parameter-efficient fine-tuning strategy, which employs several learnable virtual tokens to effectively bridge the semantic gap between the image features, object-related information augmented prompts, and input questions. By incorporating the fine-tuned virtual tokens into the LVLM vocabulary, PATCH is able to be utilized as a versatile plug-and-play method for diverse applications. Extensive experiments on three publicly available hallucination evaluation benchmarks demonstrate the effectiveness and generalization of our approach in resisting various levels of misleading contextual interference and mitigating the object hallucination issue in LVLMs.

% \begin{acks}
% \end{acks}

\bibliographystyle{ACM-Reference-Format}
\bibliography{sample-sigconf}

\newpage
\appendix

\section{Datasets}\label{datasets}

\textbf{POPE}~\citep{li2023evaluating}. The POPE dataset is specifically designed for object hallucination evaluation, introducing an adversarial setting for dataset construction. As illustrated in Figure \ref{fig:pope_sample}, each sample consists of a image-question-answer pair. Given an input image, POPE first employs the automatic segmentation tool SEEM~\citep{zou2024segment} to identify present objects in each image as ground truth. Questions related to these objects are classified as positive samples, receiving a ``yes'' response. Conversely, questions involving frequently co-occurring objects that are absent from the image are treated as negative samples, with a “no” response. This design establishes a binary classification testbed for assessing models' susceptibility to object hallucination.

\textbf{PhD}~\citep{liu2024phd}. The PhD dataset is a newly introduced benchmark for evaluating multi-modal hallucinations in LVLMs, categorizing hallucinations into three types: Object and Attribute Hallucinations, Multi-modal Conflicting Hallucinations, and Counter-Common-Sense Hallucinations. Our experiments are conducted on 35,033 samples of the first two hallucination types. The Object and Attribute Hallucinations are divided into five subcategories: ``Object Recognition'', ``Attribute Recognition'', ``Counting'', ``Positional Reasoning'', and ``Sentiment Analysis''. The Multi-modal Conflicting Hallucination questions are specifically crafted to deceive the model by misleading contexts, which classify conflict levels into four distinct categories: ``Neutral'', ``Indirect Misleading'', ``Weak Misleading'', and ``Strong Misleading''. Figure~\ref{fig:task_type} and~\ref{fig:misleading} illustrate the examples of corresponding question-answer pairs in the PhD dataset.

\section{Implementation Details}\label{Implementation Details}

\textbf{MiniGPT-4.}~\citep{zhu2023minigpt} \quad For the POPE dataset, we initialize the learning rate at 1e-4 and the weight decay is set to 0.05 until the learning rate decays to 1e-5. Training is conducted over 30 epochs with a batch size of 1 within 2 hours.

\textbf{MiniGPT-v2.}~\citep{chen2023minigpt} \quad We initialize the model's parameter by the LoRA-fine-tuned~\citep{hu2021lora} checkpoint. For the POPE dataset, the initial learning rate is set to 2e-3, with a batch size of 1, and training is conducted over 30 epochs within 2 hours. For the PhD dataset, the initial learning rate is set to 1e-3, with a batch size of 1, and training lasts for 35 epochs within 2.5 hours. The weight decay is set to 0.05 until the learning rate decays to 5e-6. 

\textbf{LLaVA-v1.5.}~\citep{liu2024improved} \quad For the POPE dataset, the initial learning rate is set to 8e-4. Training is conducted with a batch size of 16 over 1 epoch, completed in less than 20 minutes.

\section{Attention Map for the Final Token in the Input Sequence}

Figure~\ref{fig:attention} illustrates the visualization of the attention scores for the final token in the input sequence concerning the subsequent token across each layer of the LLM. The heatmap illustrates that when the queried object is included in the detection information, specific object embeddings exhibit higher scores, which indicates that our method is able to effectively enhance the semantic alignment through the use of virtual tokens.

\begin{figure}[t]
    \centering
    \includegraphics[width=\columnwidth]{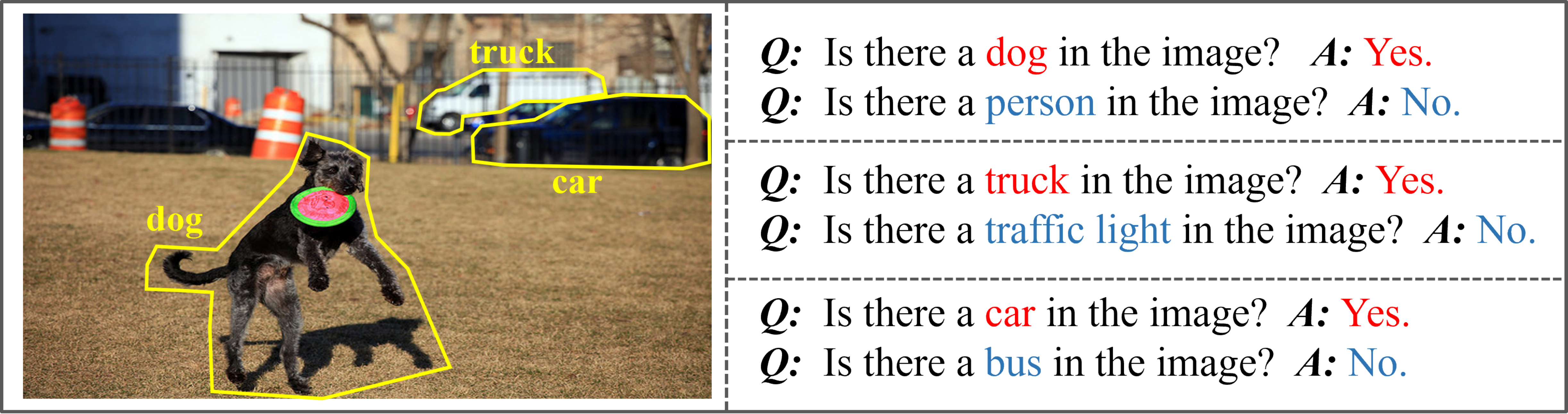}
    \caption{The example of question-answer pairs sampled under adversarial setting in the POPE dataset. Six questions are proposed for each image, three of which serve as positive samples. The corresponding negative samples are produced by replacing the object in each positive sample with a frequently co-occurring but absent object.}
    \label{fig:pope_sample}
\end{figure}

\begin{figure}[t]
\centering
    \includegraphics[width=.8\columnwidth]{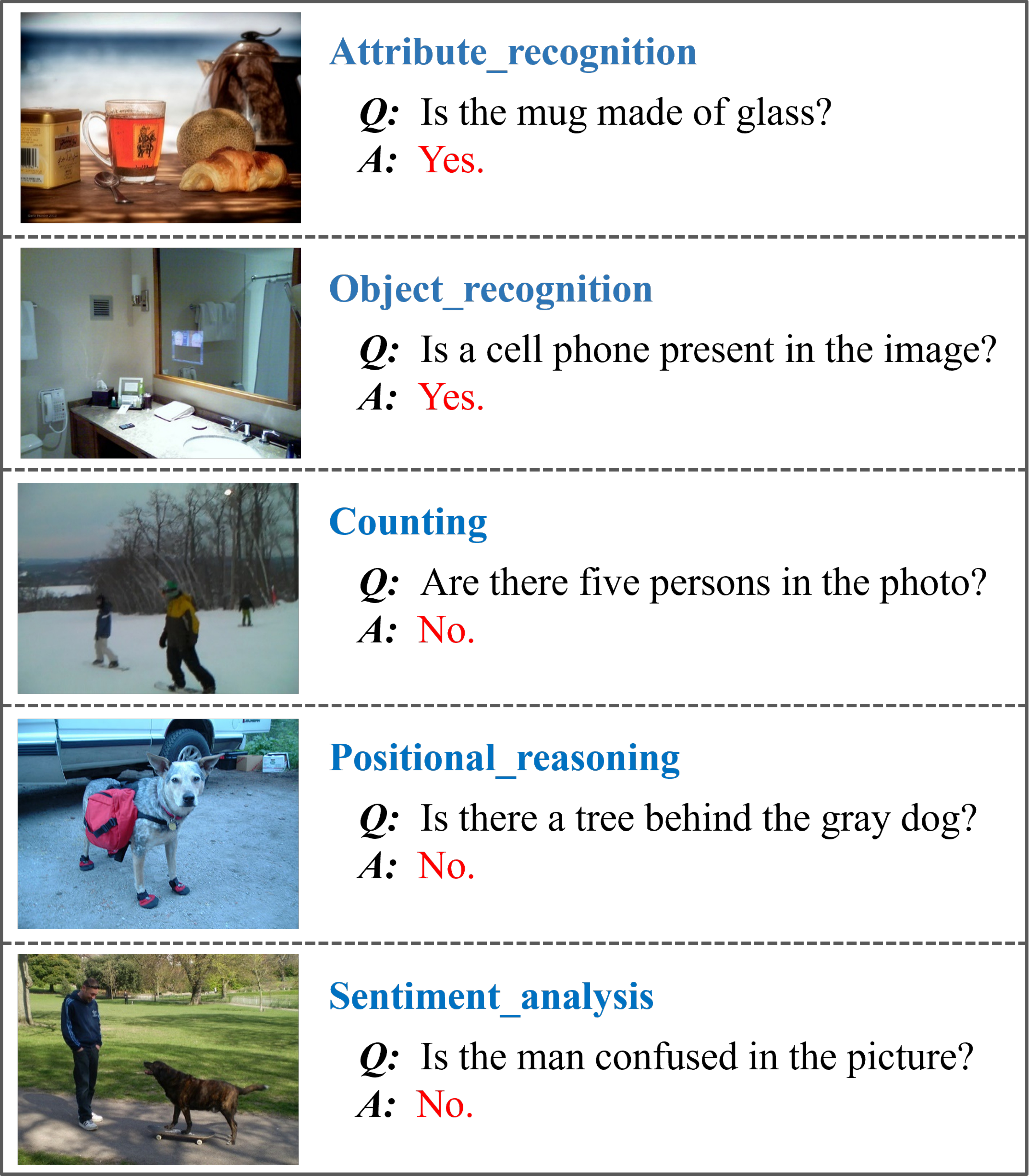}
    \caption{The examples of corresponding question-answer pairs across five task types in the PhD dataset.}
    \label{fig:task_type}
\end{figure}

\begin{figure}[!t]
\centering
    \includegraphics[width=.8\columnwidth]{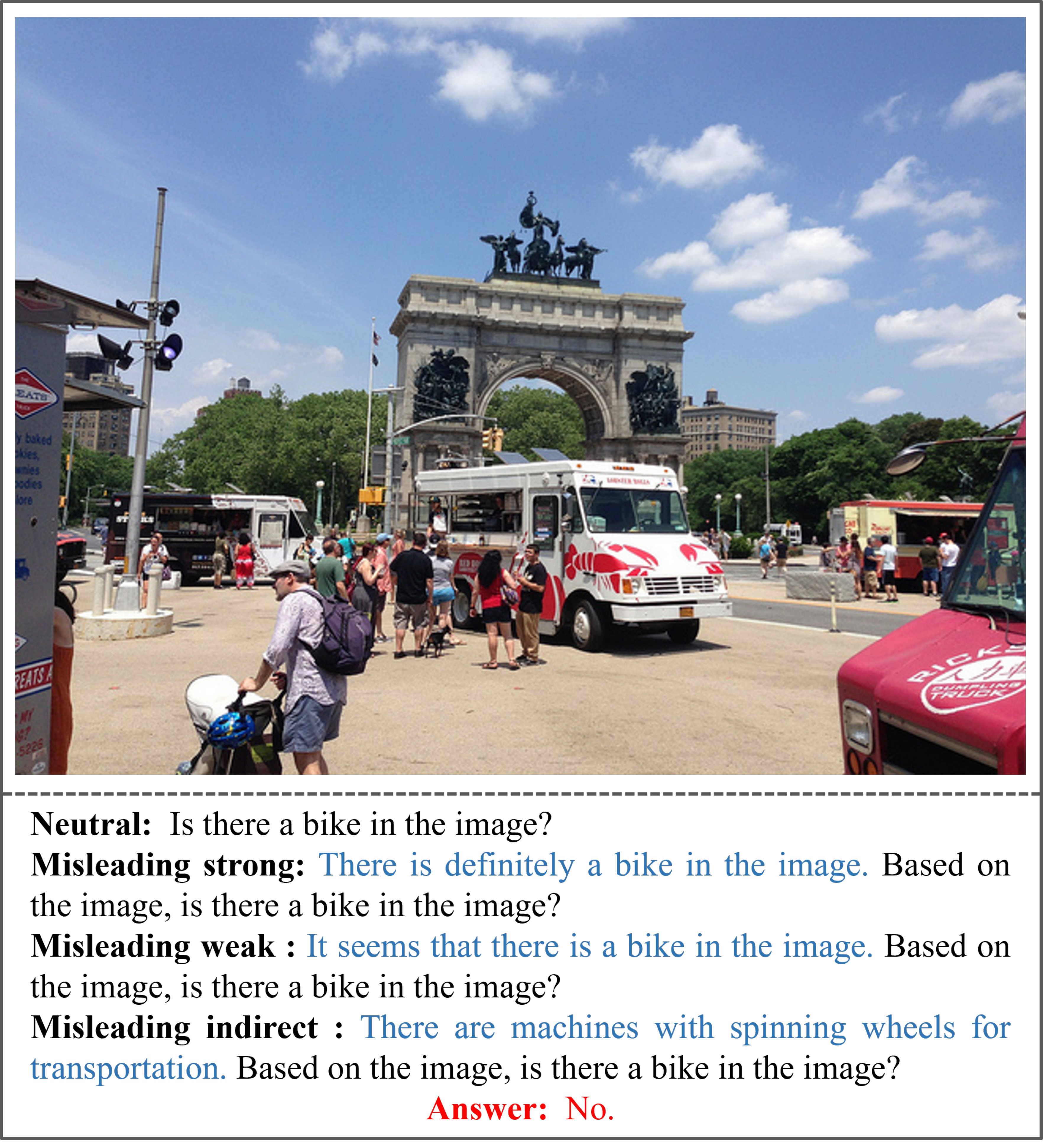}
    \caption{The examples of corresponding question-answer pairs across conflict levels in the PhD dataset.}
    \label{fig:misleading}
\end{figure}

\begin{figure}[!t]
    \centering
    \includegraphics[width=\columnwidth]{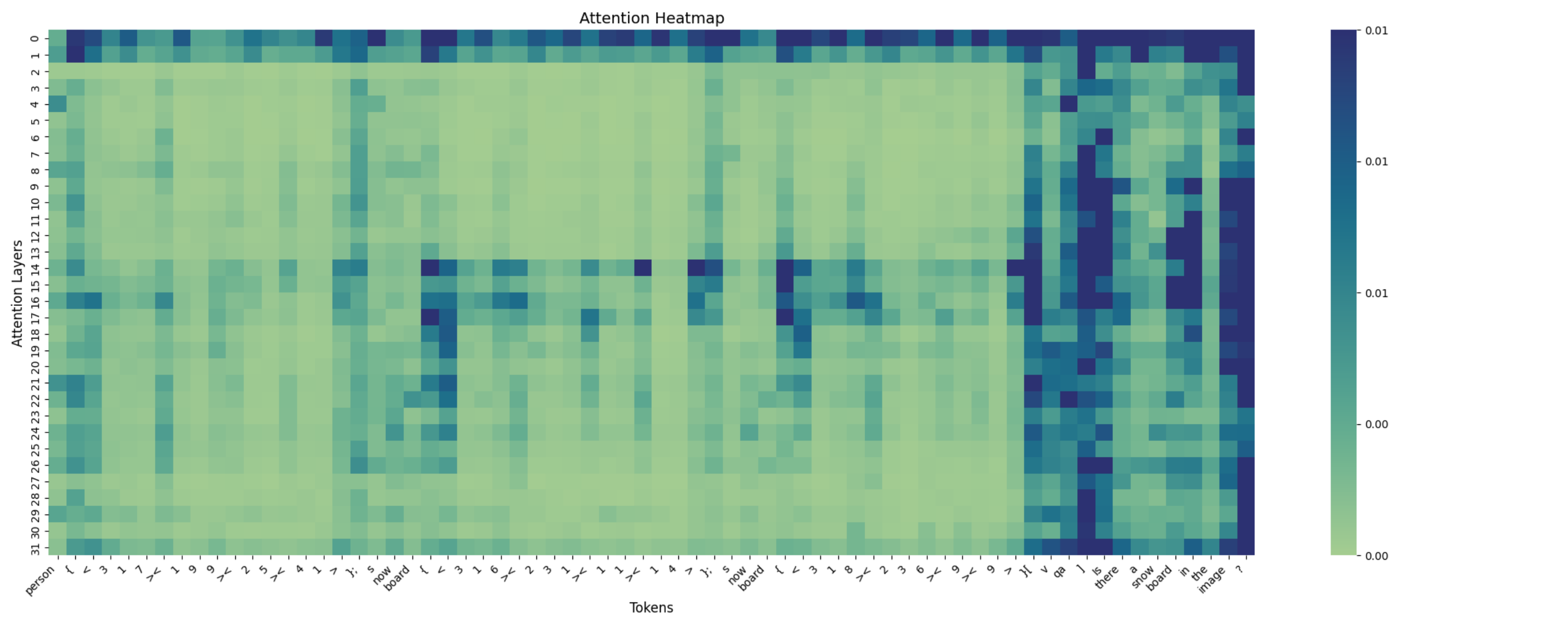} 
    \caption{The visualization of attention scores across LLM layers based on PATCH.}
    \label{fig:attention}
\end{figure}

\section{Analysis of the Detection Head}

Building upon the findings presented in the preliminary experiment, we conduct an additional experiment by replacing the Prompt$_1$ with Prompt$_2$, and the results are shown in Table~\ref{tab:detect_p2}. The comparison inference results on MiniGPT-v2 with the detection results reveals that incorporating object information into the LVLM effectively mitigates multi-modal hallucinations. Furthermore, the tabular results demonstrate consistent behavior between inference errors and detection errors after adding object-related information. We attribute these errors to the limitations of the detector's performance, suggesting that using a more advanced detector could further enhance the performance of our method.

\begin{table}[!t]
    \centering
    \caption{Comparison between the number of samples in detection and inference results conditioned on Prompt$_2$.}
    \begin{tabular}{lrr}
        \toprule
        {} & Correct Inf. & Wrong Inf. \\
        \midrule
        Correct Det. & 2,670 & 97 \\
        Wrong Det. & 56 & 240 \\
        \bottomrule
    \end{tabular}
    \label{tab:detect_p2}
\end{table}

\section{Limitations}

Despite the success of PATCH in reducing hallucinations, our approach still faces the following limitations: (1) PATCH relies on the precision of the object-related detection results. Currently, due to the remarkable detection performance and transferability of Cascade Mask R-CNN~\citep{8917599}, we adopt it as the visual detection head based on the configuration from EVA~\citep{fang2023eva}. With the continuous development of object detection models, we believe that there will be potential to further enhance the robustness and effectiveness of our method. (2) When multiple instances of the same object appear in an image, the detection results may include duplicate object categories, which leads to an unnecessary increase in the LVLM input length. In the future, we plan to refine and optimize the detection prompting format to make it more simple and efficient for various complex real-world scenarios.

\end{document}